\documentclass[accepted]{uai2024} 

\usepackage[utf8]{inputenc}
\usepackage[T1]{fontenc}

\usepackage[american]{babel}

\usepackage{natbib} 
\usepackage{mathtools}
\usepackage{amssymb}
\usepackage{bm}
\usepackage{siunitx}
\usepackage{booktabs}
\usepackage{tabu}
\usepackage{csvsimple}
\usepackage{multicol}
\usepackage{multirow}
\usepackage{tikz}
\usetikzlibrary{backgrounds,calc}
\usepackage{pgfplots}
\pgfplotsset{compat=1.5}
\pgfplotsset{
    cycle list={t_blue\\t_red\\t_green\\t_yellow\\},
}
\usepgfplotslibrary{fillbetween}
\usepackage{intcalc}
\usepackage{graphicx}

\usepackage{hyperref}
\hypersetup{
    plainpages=false,           
    pdfborder={0 0 0},          
    breaklinks=true,            
    bookmarksnumbered=true,     %
    bookmarksopen=true,         %
	hypertexnames=true,         %
}

\usepackage[capitalise,nameinlink]{cleveref}
\crefname{ax}{axiom}{axioms}
\usepackage{acro}
\acsetup{format/first-long=\slshape} 
\acsetup{barriers/use=true}
\acsetup{barriers/reset=true}
\acsetup{single}

\usepackage{colortbl}
\definecolor{t_gray}{HTML}{888888}
\definecolor{t_blue}{HTML}{355fb3}
\definecolor{t_red}{HTML}{b33535}
\definecolor{t_green}{HTML}{3bb335}
\definecolor{t_yellow}{HTML}{b39735}
\definecolor{t_darkgray}{HTML}{454545}
\definecolor{t_darkblue}{HTML}{1e3666}
\definecolor{t_darkgreen}{HTML}{22661e}
\definecolor{t_darkred}{HTML}{661e1e}
\definecolor{t_darkyellow}{HTML}{66571e}
\definecolor{t_lightblue}{HTML}{8ea7d7}
\definecolor{t_lightred}{HTML}{dc8989}
\definecolor{t_lightgreen}{HTML}{8ddc89}

\newcommand{\dac}[3]{\DeclareAcronym{#1}{short = #2, long = #3}}

\DeclareMathOperator*{\argmin}{arg\,min}

\newcommand{\bftab}{\fontseries{b}\selectfont}
\newcommand*{\condbold}[3][]{\ifthenelse{\equal{#2}{1}#1}{{\bftab #3}}{#3}}
\newcommand*{\rawres}[3]{\ifthenelse{\equal{#1}{}}{-}{\condbold{#3}{#1}}}
\pgfplotscreateplotcyclelist{accrejPlotColors}{%
t_green\\%
t_blue\\%
t_red\\%
}
\newcommand*{\addSE}[3]{
    \addplot[name path=#2upper, draw=none] table[x=p, y expr=\thisrow{#2Mean}+\thisrow{#2SE}, col sep=comma] {#1};
    \addplot[name path=#2lower, draw=none] table[x=p, y expr=\thisrow{#2Mean}-\thisrow{#2SE}, col sep=comma] {#1};
    \addplot[fill=#3, fill opacity=0.2] fill between[of=#2upper and #2lower];
}

\dac{ml}{ML}{machine learning}
\dac{nlp}{NLP}{natural language processing}
\dac{nn}{NN}{neural network}

\dac{mlp}{MLP}{multilayer perceptron}
\dac{rnn}{RNN}{recurrent neural network}
\dac{svm}{SVM}{support vector machine}
\dac{cnn}{CNN}{convolutional neural network}
\dac{gnn}{GNN}{graph neural network}
\dac{gcn}{GCN}{Graph Convolutional Network}
\dac{gin}{GIN}{Graph Isomorphism Network}
\dac{gpn}{GPN}{Graph Posterior Network}
\dac{lop}{LOP}{linear opinion pooling}
\dac{lopgpn}{LOP-GPN}{linear opinion pooled graph posterior network}
\dac{au}{AU}{aleatoric uncertainty}
\dac{eu}{EU}{epistemic uncertainty}
\dac{tu}{TU}{total uncertainty}
\dac{uq}{UQ}{uncertainty quantification}
\dac{appnp}{APPNP}{approximate personalized propagation of neural predictions}
\dac{postnet}{PostNet}{Posterior Network}
\dac{ce}{CE}{cross-entropy}
\dac{uce}{UCE}{uncertain cross-entropy}
\dac{ppr}{PPR}{personalized page-rank}
\dac{arc}{ARC}{accuracy-rejection curve}
\dac{ood}{OOD}{out-of-distribution}
\dac{id}{ID}{in-distribution}
\dac{auroc}{AUC-ROC}{area under the receiver operating characteristic curve}

\title{Linear Opinion Pooling for Uncertainty Quantification on Graphs}

\author[1]{\href{mailto:<clemens.damke@ifi.lmu.de>?Subject=Your UAI 2024 paper}{Clemens Damke}{}}
\author[1,2]{\href{mailto:<eyke@ifi.lmu.de>?Subject=Your UAI 2024 paper}{Eyke Hüllermeier}{}}
\affil[1]{%
    Institute of Informatics\\
    LMU Munich\\
    Germany
}
\affil[2]{%
    Munich Center for Machine Learning (MCML)
}

\begin{document}
\maketitle

\begin{abstract}
We address the problem of uncertainty quantification for graph-structured data, or, more specifically, the problem to quantify the predictive uncertainty in (semi-supervised) node classification. Key questions in this regard concern the distinction  between two different types of uncertainty, aleatoric and epistemic, and how to support uncertainty quantification by leveraging the structural information provided by the graph topology. Challenging assumptions and postulates of state-of-the-art methods, we propose a novel approach that represents (epistemic) uncertainty in terms of mixtures of Dirichlet distributions and refers to the established principle of linear opinion pooling for propagating information between neighbored nodes in the graph. The effectiveness of this approach is demonstrated in a series of experiments on a variety of graph-structured datasets.
\end{abstract}

\section{Introduction}\label{sec:intro}

Quantifying the uncertainty of predictions made by machine learning models is critical for applications where safety is important and mistakes can be costly.
When assessing the uncertainty of a model's prediction, it is common and often useful to distinguish between two types of uncertainty: 
\Ac{au} arises from the stochasticity inherent to the data-generating process, and cannot be reduced by sampling additional data.
For example, when tossing a fair coin, the outcome is uncertain, and this uncertainty is of purely aleatoric nature.
\Ac{eu}, on the other hand, is due to a lack of knowledge about the data-generating process; assuming that an appropriate model class is chosen, \ac{eu} can be reduced by collecting more data and vanishes in the limit of infinite data~\citep{hullermeier2021}.
For example, the lack of knowledge about the bias of a coin is of epistemic nature, and it increases the (total) uncertainty about the outcome of a coin toss. This uncertainty, however, can be reduced by tossing the coin repeatedly and estimating the bias from the outcomes.

In the context of graph-structured data, \ac{uq} is particularly challenging due to the structural information as an additional contributing factor to the uncertainty.
In this paper, we will focus specifically on the problem of \ac{uq} for (semi-supervised) node classification.
Applications of this problem include, for example, the classification of documents in citation networks~\citep{sen2008,bojchevski2018}, or the classification of users or posts in social networks~\citep{shu2017}.

Recently, \acp{gpn} have been proposed as a principled approach to \ac{uq} on graphs~\citep{stadler2021}.
The \ac{gpn} model combines \acp{postnet}~\citep{charpentier2020} with the \ac{appnp} node classification model~\citep{gasteiger2018}.
This combination is motivated by three axioms on how the structural information in a graph should affect the uncertainty of a model's predictions.
One of those axioms states that the aleatoric entropy of nodes with conflicting neighbors should be high.

In this paper, we discuss the validity of this assumption and situations in which it does not hold.
To address those situations, we propose a novel approach to \acl{uq} on graphs based on the idea of \ac{lop} from the field of decision and risk analysis~\citep{clemen2007,stone1961}.
Our approach, which we refer to as \ac{lopgpn}, uses a mixture of Dirichlet distributions to model the uncertainty of a node's label.
We demonstrate the effectiveness of our approach in a series of experiments on a variety of graph-structured datasets, showing that it outperforms existing methods in terms of both predictive accuracy and uncertainty quantification.

The remainder of this paper is organized as follows.
In \cref{sec:um} we give an overview of commonly used measures for \ac{uq}.
In \cref{sec:uq}, we review how \acp{gpn} use those measures to quantify their predictive uncertainty and then discuss the validity of this approach and describe its problems.
\Cref{sec:lop} presents the \acp{lopgpn} approach that addresses the problems described in the previous section.
In \cref{sec:eval}, we compare our approach with the original \ac{gpn} model and other baseline models.
Finally, \cref{sec:conclusion} concludes the paper and outlines directions for future work.

\section{Uncertainty Measures}\label{sec:um}

In the literature on \ac{uq}, there are different notions of what formally constitutes uncertainty.
Depending on the desired properties of the uncertainty measure, different notions may be more or less suitable.
We consider two ways to assess the suitability of a measure of uncertainty:
\begin{enumerate}
  \item Its adherence to a set of axioms~\citep{pal1993,bronevich2008,wimmer2023,sale2023a}.
    \item Its performance on a predictive task, such as outlier detection~\citep{charpentier2020}.
\end{enumerate}
Since we focus on \ac{uq} in the node classification setting, we provide a brief overview of the most common notions of uncertainty in the context of classification tasks.

\subsection{Entropy-based Uncertainty Measures}\label{sec:um:entropy}

One of the most common ways to represent predictive uncertainty of a $K$-class classifier is through the use of a second-order probability distribution $Q$, i.e., a distribution on the probability distributions $\theta = (\theta_1, \ldots , \theta_K)\in \Delta_K$, where $\Delta_K$ is the unit $(K-1)$-simplex and $\theta_k$ denotes the probability of the $k^{th}$ class.
Thus, the true distribution on the $K$ classes is considered as a random variable $\Theta \sim Q$.
Given a second-order distribution $Q$, we denote its expectation by
\begin{equation}
    \bar{\theta} \coloneqq \mathbb{E}_{Q}[\Theta] = \int_{\Delta_K} \theta \, \mathrm{d}Q(\theta) \, . \label{eq:q-exp}
\end{equation}
The \acf{tu} (about the outcome $Y$, i.e., the class eventually observed) can be quantified by the Shannon entropy of $\bar{\theta}$, i.e.,
\begin{equation}
    \mathrm{TU}(Q) \coloneqq H(\mathbb{E}_{Q}[\theta]) = -\sum_{k=1}^{K} \bar{\theta}_k \log \bar{\theta}_k . \label{eq:tu}
\end{equation}
Further, a decomposition of this uncertainty into an aleatoric and an epistemic part can be achieved on the basis of a well-known result from information theory, stating that entropy is the sum of conditional entropy and mutual information~\citep{kendall2017,depeweg2018}.
This result suggests to quantify \acf{au} as conditional entropy (of the outcome $Y$ given the first-order distribution $\Theta$):
\begin{equation}
    \mathrm{AU}(Q) \coloneqq \mathbb{E}_{Q} \left[ H(\Theta) \right] = -\smashoperator{\int_{\Delta_K}} \sum_{k=1}^{K} \theta_k \log \theta_k \, \mathrm{d}Q(\theta) \, . \label{eq:au}
\end{equation}
Moreover, the \acf{eu} is then given by the difference between \ac{tu} and \ac{au}, i.e.,
\begin{align}
    \mathrm{EU}(Q) & \coloneqq \mathrm{TU}(Q) - \mathrm{AU}(Q) \label{eq:eu} \\
    & = I(Y; \Theta) = \mathbb{E}_G[D_{\mathrm{KL}}(\Theta \| \bar{\theta})] \, ,  \nonumber
\end{align}
where $I(\cdot; \cdot)$ denotes mutual information and $D_{\mathrm{KL}}(\cdot \| \cdot)$ the Kullback-Leibler divergence.

\begin{figure}[t]
    \centering
    \includegraphics[width=\linewidth]{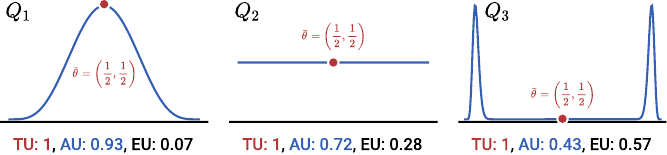}
    \caption{%
        \acs*{tu}, \acs*{au}, and \acs*{eu} for three second-order distributions on $\Delta_2$,
        namely, $Q_1 = \mathrm{Beta}(5,5)$, $Q_2 = \mathcal{U}[0,1]$, and $Q_3 = \frac{1}{2}\mathrm{Beta}(100,10) + \frac{1}{2}\mathrm{Beta}(10,100)$.%
    }\label{fig:entropy-uncertainty}
\end{figure}
\Cref{fig:entropy-uncertainty} gives an intuition for the behavior of this additive decomposition of uncertainty (for $K=2$).
It shows how the \ac{au} goes down for second-order-distributions which are more concentrated around degenerate categorical distributions.
As described by \citet{wimmer2023}, there are however situations in which this decomposition is less plausible.
For example, $Q_2$ and $Q_3$ in \cref{fig:entropy-uncertainty} have the same \ac{tu}, despite the fact that $Q_3$ arguably expresses more knowledge about the true data-generating distribution $\theta^*$ than $Q_2$.
This raises the question, whether an additive decomposition of \ac{tu} into \ac{au} and \ac{eu} is reasonable at all.

Instead of expressing \ac{eu} in terms of the mutual information between $Y$ and $\Theta$, \citet{malinin2018} and \citet{kotelevskii2023} propose to express \ac{eu} via the differential entropy of the second-order distribution $Q$, i.e.,
\begin{equation}
    \mathrm{EU}_{\textrm{SO}}(Q) \coloneqq H(Q) = -\int_{\Delta_K} \log Q(\theta) \, \mathrm{d}Q(\theta). \label{eq:eu-so-entropy}
\end{equation}
In the following, we will refer to this notion of \ac{eu} as \emph{second-order epistemic uncertainty}.
Note that this notion of \ac{eu} is not without controversy, as the differential entropy can be negative, which is forbidden in some axiomatic characterizations of uncertainty, which use $0$ to represent a state of no uncertainty~\citep{wimmer2023}.
Apart from the entropy-based measures we just described, uncertainty is also often quantified in terms of other concentration measures, such as variance~\citep{sale2023a,duan2024}, confidence or Dirichlet pseudo-counts.
We will now briefly review the latter two notions of uncertainty.

\subsection{Least-confidence- and Count-based Uncertainty Measures}\label{sec:um:confidence}

Given a second-order distribution $Q$, a different notion of uncertainty is provided by the so-called \emph{least-confidence} of the expected distribution $\bar{\theta}$, defined as
\begin{equation}
    \mathrm{LConf}(Q) \coloneqq 1 - \max_k \bar{\theta}_k \, . \label{eq:lconf}
\end{equation}
Note the similarity of this measure to the \ac{tu} measure in \cref{eq:tu}; $\mathrm{LConf}(Q)$ can therefore be seen as a measure of \emph{total} uncertainty.
However, in the literature this measure is also used as a proxy for \emph{aleatoric} uncertainty~\citep{charpentier2020}; we will come back to this point in \cref{sec:uq}.

Finally, if $Q$ is described by a Dirichlet distribution $\mathrm{Dir}(\bm{\alpha})$, where $\bm{\alpha} = (\alpha_1, \dots, \alpha_K)$ is a vector of pseudo-counts,
the sum $\alpha_0 = \sum_{k=1}^K \alpha_k$ describes how concentrated $Q$ is around the expected distribution $\bar{\theta}$.
Note that the concentration of $Q$ is similarly captured by its differential entropy, as described in \cref{eq:eu-so-entropy}, i.e., $\mathrm{EU}_{\textrm{SO}}(\mathrm{Dir}(\bm{\alpha}))$ goes down as $\alpha_0$ grows.
The \ac{eu} of a Dirichlet distribution $Q$ can therefore be quantified by $\mathrm{EU}_{\textrm{PC}}(Q) = -\alpha_0$.
We will refer to this notion of \ac{eu} as \emph{pseudo-count-based epistemic uncertainty}~\citep{charpentier2020,huseljic2021,kopetzki2021}.

\section{Uncertainty Quantification}\label{sec:uq}

As just described, there are different ways to formalize uncertainty, and the choice of an uncertainty measure depends on the properties it is supposed to fulfill.
In the context of graphs, there is an additional factor that contributes to uncertainty and needs to be formalized, too, namely the structural information.
\Citet{stadler2021} propose an axiomatic approach to account for this structure-induced uncertainty, which they call \acf{gpn}.
As mentioned in the introduction, \acp{gpn} are essentially a combination of \acp{postnet}~\citep{charpentier2020} and the \ac{appnp} node classification model~\citep{gasteiger2018}.
We begin with a brief review of the \acp{postnet} and \acp{gpn}, and then discuss the validity of the axioms on which the \ac{uq} estimates of \acp{gpn} are essentially based.

\subsection{Posterior Networks}\label{sec:uq:postnet}

A \ac{postnet} is a so-called \emph{evidential deep learning} classification model~\citep{sensoy2018}, i.e., it quantifies predictive uncertainty via a second-order distribution $Q$, which is learned via a \emph{second-order loss function} $L_2$.
A standard (first-order) loss function $L_1 : \Delta_K \times \mathcal{Y} \to \mathbb{R}$ takes a predicted first-order distribution $\hat{\theta} \in \Delta_K$ and an observed ground-truth label $y \in \mathcal{Y}$ as input (where $\mathcal{Y}$ denotes the set of classes); the \acf{ce} loss is a common example of such a first-order loss function.
Similarly, a second-order loss $L_2$ takes a second-order distribution $Q$, i.e., a distribution over $\Delta_K$, as input, to which it again assigns a loss in light of an observed label $y \in \mathcal{Y}$.
\Ac{postnet} uses the so-called \acf{uce} loss~\citep{bilos2019}, which is defined as
\begin{align}
    L_2(Q, y) & \coloneqq \mathbb{E}_{Q} \left[ \mathrm{CE}(\Theta, y) \right]  \label{eq:uce}  \\
    & = -\int_{\Delta_K} \log P(y\, |\, \theta) \, \mathrm{d}Q(\theta) \, . \nonumber
\end{align}
As shown by \citet{bengs2022}, directly minimizing a second order loss, like the \ac{uce} loss, is problematic, since the minimum is reached if $Q$ is a Dirac measure that puts all probability mass on $\theta^* = {\argmin}_{\theta \in \Delta_K} \mathrm{CE}(\theta, y)$.
For all notions of \ac{eu} we discussed in \cref{sec:um}, the optimal $Q^*$ will therefore have no \ac{eu}, i.e., $\mathrm{EU} = 0$~(\cref{eq:eu}) and $\mathrm{EU}_{\textrm{SO}} = \mathrm{EU}_{\textrm{PC}} = -\infty$~(\cref{eq:eu-so-entropy}).
This problem is commonly addressed by adding a \emph{regularization term}, typically the differential entropy of $Q$, in the second-order loss function, which encourages $Q$ to be more spread out.
Whether one can obtain a faithful representation of \acl{eu} has been generally questioned by \cite{bengs2023}.
One should therefore be cautious when interpreting the \ac{eu} estimates of evidential deep learning models, such as \ac{postnet}.
We will not attempt to interpret uncertainty estimates in a quantitative manner, but rather focus on the question of whether they are qualitatively meaningful, e.g., by considering whether anomalous or noisy instances can be identified via their uncertainty.

\Ac{postnet} models the second-order distribution $Q$ as a Dirichlet distribution $\mathrm{Dir}(\mathbf{\alpha})$, where $\mathbf{\alpha} = (\alpha_1, \dots, \alpha_K)$ is a vector of pseudo-counts.
The predicted pseudo-counts $\alpha_k$ for a given instance $\mathbf{x}^{(i)} \in \mathcal{X}$ are defined as
\begin{equation}
    \alpha_k = 1 + N \cdot P \left(\mathbf{z}^{(i)}\,|\, y^{(i)} = k \right) \cdot P \left(y^{(i)} = k \right) ,
\end{equation}
where $\mathbf{z}^{(i)} = f(\mathbf{x}^{(i)}) \in \mathbb{R}^H$ is a latent neural network embedding of $\mathbf{x}^{(i)}$ and $N \in \mathbb{R}$ a so-called \emph{certainty budget}, determining the highest attainable pseudo-count for a given instance.
The class-conditional probability $P(\mathbf{z}^{(i)}\,|\, k)$ by a normalizing flow model for the class $k$ estimates the density of the instance.
Overall, the \ac{postnet} model therefore consists of a neural network encoder model $f$ and $K$ normalizing flow models, one for each class.

\subsection{Graph Posterior Networks}\label{sec:uq:gpn}

Next, we describe how the \acf{gpn} approach extends \ac{postnet} to the node classification setting for graphs.
We denote a graph as $G \coloneqq (\mathcal{V}, \mathcal{E})$, where $\mathcal{V}$ is a set of $N \coloneqq |\mathcal{V}|$ nodes and $\mathcal{E} \subseteq \mathcal{V}^2$ the set of edges.
The adjacency matrix of $G$ is denoted by $\mathbf{A} = (A_{i,j}) \in {\{0,1\}}^{N \times N}$, where $A_{i,j} = 1$ iff $(v_i, v_j) \in \mathcal{E}$.
For simplicity we also assume that $G$ is undirected, i.e., that $A$ is symmetric.
For each node $v^{(i)} \in \mathcal{V}$ we have a feature vector $\mathbf{x}^{(i)} \in \mathbb{R}^D$ and a label $y^{(i)} \in \mathcal{Y}$.
The goal of the node classification task is to predict the label of each node in $\mathcal{V}$, given the graph structure and the node features.

\Acp{gpn} classify the nodes of a given graph by first making a prediction for each node $v^{(i)}$ solely based on its features $\mathbf{x}^{(i)}$ using a standard \ac{postnet} model, i.e., without considering the graph structure.
The predicted feature-based pseudo-count vectors $\mathbf{\alpha}^{\mathrm{ft},(i)}$ for each vertex $v^{(i)}$ are then dispersed through the graph via a \ac{ppr} matrix $\Pi^{\mathrm{PPR}} \in \mathbb{R}^{N \times N}$ as follows:
\begin{align}
    \mathbf{\alpha}^{\mathrm{agg},(i)} &\coloneqq \sum_{v^{(j)} \in \mathcal{V}} \Pi^{\mathrm{PPR}}_{i,j} \mathbf{\alpha}^{\mathrm{ft},(j)} \label{eq:gpn-alpha} \\
    \text{where } \Pi^{\mathrm{PPR}} &\coloneqq {\left(\varepsilon \mathbf{I} + (1 - \varepsilon) \mathbf{\hat{A}}\right)}^L \label{eq:gpn-ppr}
\end{align}
Here, $\mathbf{I}$ is the identity matrix, $\varepsilon \in (0,1]$ the so-called \emph{teleport probability}, and $\mathbf{\hat{A}} \coloneqq \mathbf{A} \mathbf{D}^{-1}$ the normalized (random-walk) adjacency matrix, with $\mathbf{D} \coloneqq \mathrm{diag}(\mathbf{A} \mathbf{1})$ being the degree matrix of $G$.
For large $L$, $\Pi^{\mathrm{PPR}}$ approximates the personalized page-rank matrix of the graph via power iteration.
\Citet{gasteiger2018} proposed this page-rank inspired information dispersion scheme for the node classification task, which they refer to as \ac{appnp}.
The main difference between \ac{appnp} and \ac{gpn} is that \ac{appnp} disperses (first-order) class probability vectors $\theta^{\mathrm{ft},(i)}$ for each node $v^{(i)}$, whereas \ac{gpn} disperses pseudo-count vectors $\mathbf{\alpha}^{\mathrm{ft},(i)}$. The latter correspond to second-order parameters, but are also in direct correspondence to zero-order (pseudo-)data. Broadly speaking, \ac{appnp} disperses first-order information, whereas \ac{gpn} disperses zero-order information.

To justify this pseudo-count dispersion scheme, \citet{stadler2021} propose three axioms on how the structural information in a graph should affect the uncertainty of a model's predictions:
\begin{enumerate}[label=\textbf{A\arabic*}]
    \item A node's prediction should only depend on its own features in the absence of network effects.
        A node with features more different from training features should have a higher uncertainty. \label[ax]{ax:gpn-ft}
    \item All else being equal, if a node $v^{(i)}$ has a lower epistemic uncertainty than its neighbors in the absence of network effects, the neighbors' predictions should become less epistemically uncertain in the presence of network effects. \label[ax]{ax:gpn-eu}
    \item All else being equal, if a node $v^{(i)}$ has a higher aleatoric uncertainty than its neighbors in the absence of network effects, the neighbors' predictions should become more aleatorically uncertain in the presence of network effects.
        Further, the aleatoric uncertainty of a node in the presence of network effects should be higher if the predictions of its neighbors in the absence of network effects are more conflicting. \label[ax]{ax:gpn-au}
\end{enumerate}
To show the validity of those axioms, \citet{stadler2021} define \ac{au} to be the least-confidence (\cref{eq:lconf}) and \ac{eu} as the sum of the pseudo-counts.
Using those definitions, the validity of the axioms follows from the fact that $\mathbf{\alpha}^{\mathrm{agg},(i)}$ is effectively a weighted average of the pseudo-counts of the (indirect) neighbors of $v^{(i)}$, with high weights for close neighbors and low weights for more distant ones.

\subsection{Validity of the GPN Axioms}\label{sec:uq:critique}

\begin{figure}[t]
    \centering
    \includegraphics[width=\linewidth]{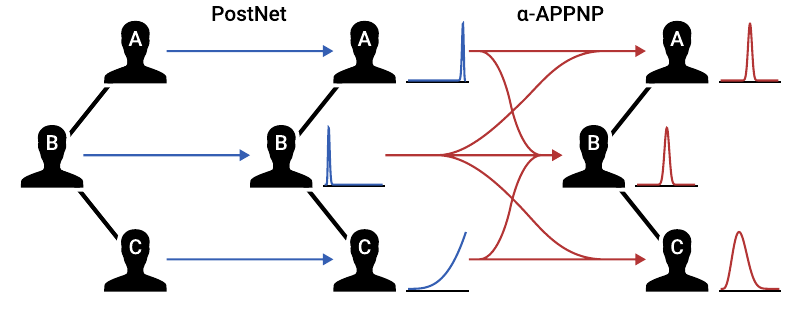}
    \caption{Illustration of how \acs*{gpn} aggregates the two conflicting predictions for \textsf{A} and \textsf{B} with low \acs*{au} and low \acs*{eu} into predictions with high \acs*{au} and low \acs*{eu}.}\label{fig:gpn-aggregation}
\end{figure}
The axioms that motivate the aggregation of pseudo-counts in \acp{gpn} are based on two assumptions which may not always hold, namely, \emph{network homophily} and the \emph{irreducibility of conflicts}.

First, \emph{network homophily} refers to the assumption that an edge implies that the connected nodes are similar in some way; more specifically, in the context of \acp{gpn} that connected nodes should have similar second-order distributions, and thereby similar predictive uncertainties.
This is a common assumption which is shared by many \ac{gnn} architectures, based on the idea of repeatedly summing or averaging the features of each node's neighbors~\citep{kipf2017,xu2018}.
As already remarked by \citet{stadler2021}, non-homophilic graphs are not properly dealt with by \acp{gpn}, nor by other \ac{gnn} architectures in general~\citep{zhu2020}.
Nonetheless, since edges are typically used to represent some form of similarity, the homophily assumption is often reasonable.

Second, we define the \emph{irreducibility of conflicts} as the assumption that conflicting predictions cannot be resolved by aggregating the predictions of the conflicting nodes.
\Cref{fig:gpn-aggregation} illustrates the implications of this assumption for the binary node classification; there, without network effects, node \textsf{A} is very confident that its probability of belonging to the positive class is high, whereas node \textsf{B} is very confident that its probability of belonging to that class is low.
Thus, both nodes make conflicting predictions, while both having a low \ac{au} and a low \ac{eu}.
Due to the homophily assumption, a consensus has to be found between the two conflicting predictions.
As described in \cref{ax:gpn-au}, a \ac{gpn} will do this by increasing the \ac{au} of the aggregated prediction, while keeping the \ac{eu} low.
\Citet{stadler2021} argue that this is reasonable, because such a conflict is inherently irreducible and should therefore be reflected in the \acl{au} of the aggregated prediction.

To assess whether the irreducibility assumption is indeed reasonable, one has to clarify what \emph{irreducibility} is actually supposed to mean.
As mentioned in the introduction, irreducibility in the context of \ac{uq} refers to uncertainty that cannot be reduced by additional information, which, in a machine learning context, essentially means by sampling additional data~\citep{hullermeier2021}.
In the context of node classification, the data points are nodes; thus given a sample graph $G_N = (\mathcal{V}_N, \mathcal{E}_N)$ with $N$ vertices, increasing the sample size corresponds to sampling a graph $G_M = (\mathcal{V}_M, \mathcal{E}_M)$ with $M > N$ nodes from an assumed underlying data-generating distribution $P_\mathcal{G}$ over all graphs $\mathcal{G}$, such that $G_N$ is a subgraph of $G_M$.
The question of whether a conflict between a node $v^{(i)}$ and its neighbor $v^{(j)}$ is irreducible then becomes the question of whether the conflict persists in the limit of $M \to \infty$.
Let $\mathcal{N}_M(v^{(i)})$ be the set of neighbors of $v^{(i)}$ in $G_M$.
Assuming homophily, each node $v^{(\ell)}$ that is added to $\mathcal{N}_M(v^{(i)})$ should be \emph{similar} to $v^{(i)}$ with high probability.
Depending on the data-generating distribution $P_\mathcal{G}$, there are two possible scenarios:
\begin{enumerate}
    \item The neighborhood of $v^{(i)}$ does not grow with the sample size, i.e., $\mathbb{E}[|\mathcal{N}_M(v^{(i)})|] \in \mathcal{O}(1)$ as $M \to \infty$.
    \item The neighborhood of $v^{(i)}$ grows with the sample size, i.e., $\mathbb{E}[|\mathcal{N}_M(v^{(i)})|] \to \infty$ as $M \to \infty$.
\end{enumerate}
In the first case, the conflict between $v^{(i)}$ and $v^{(j)}$ is indeed irreducible, as no additional data can be sampled to resolve the conflict.
In this situation, \cref{ax:gpn-au} of \ac{gpn} is reasonable, the irreducible uncertainty, i.e., \ac{au}, should increase with conflicting predictions.
However, in the second case, the conflict is reducible, as the conflict will eventually be resolved by the addition of more similar nodes to the neighborhood of $v^{(i)}$, which will outweigh the conflicting node $v^{(j)}$.
In this situation, the conflict resolution approach of \ac{gpn} is not reasonable; \ac{au} should not go up, instead the reducible uncertainty, i.e., \ac{eu}, should increase.

We argue that the second case is more common in practical node classification tasks.
The Barabási-Albert model~\citep{barabasi1999} is a popular scale-free model, which describes the growth behavior of many real-world graphs, such as the World Wide Web, social networks, or citation networks~\citep{albert2002,redner1998,wang2008}.
In this model, the expected degree of the $N$-th sampled node $v^{(i)}$ after $M-N$ additional nodes have been sampled is equal to $|\mathcal{N}_N(v^{(i)})| \cdot \sqrt{\frac{M}{N}}$.
Thus, for $M \to \infty$, the expected neighborhood size of a node goes to infinity.

Examples for domains in which the neighborhood sizes do not grow with the size of a graph are molecular graphs or lattice graphs, such as 3D models or images, which can be interpreted as grids of pixels.
In those domains, node-level classification tasks are however less common, as one is typically interested in the classification of entire graphs, e.g., whether a given molecule is toxic or not.

To conclude, we argue that the axiomatic motivation for \acp{gpn} is oftentimes inappropriate.
Therefore, we propose a different approach to \ac{uq} for node classification which does not assume the irreducibility of conflicts from \cref{ax:gpn-au}.

\section{LOP-GPN}\label{sec:lop}

\begin{figure}[t]
    \centering
    \includegraphics[width=\linewidth]{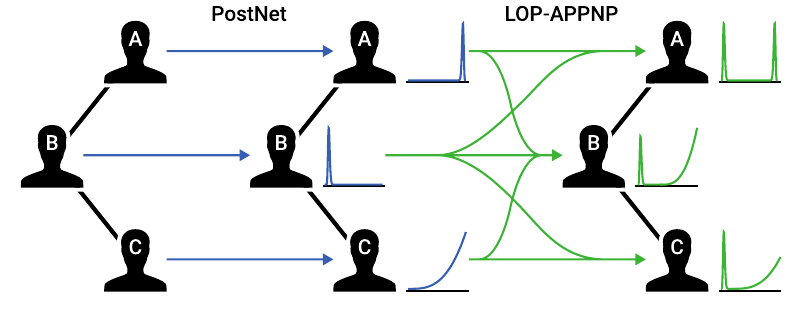}
    \caption{Illustration of how \acs*{lopgpn} preserves the \acs*{au} of conflicting predictions.}\label{fig:lop-gpn-aggregation}
\end{figure}
A \acf{lopgpn} is a variant of the standard \ac{gpn} model, which uses a mixture of Dirichlet distributions instead of a single Dirichlet distribution to model the uncertainty of a node's class in the presence of network effects.
This choice is motivated by the idea that each node $v^{(i)}$ in a given node classification task can be interpreted as a decision maker, which has to assign a label $y^{(i)}$ to itself based on its own features $\mathbf{x}^{(i)}$ and the features of its neighbors.
Using this interpretation, in the \ac{gpn} architecture, each decision maker $v^{(i)}$ first makes a prediction based on its own features, producing a Dirichlet distribution $\mathrm{Dir}(\bm{\alpha}^{\mathrm{ft},(i)})$, and then aggregates this distribution with those produced by its direct and indirect neighbors.
In the standard \acp{gpn} model this aggregate is again a Dirichlet distribution $\mathrm{Dir}(\bm{\alpha}^{\mathrm{agg},(i)})$ (see \cref{eq:gpn-alpha}).

The question of how to aggregate the decisions or opinions of a set of experts is a well-known problem in the field of decision and risk analysis~\citep{clemen2007}.
One of the most common aggregation approaches in this field is the called \acf{lop}~\citep{stone1961}.
\Ac{lop} simply aggregates distributions by taking a weighted average of their densities, resulting in a mixture of the original distributions.
Combining \ac{lop} with \ac{appnp}, the aggregated probability density of a node $v^{(i)}$ is given by
\begin{align}
    Q^{\mathrm{agg},(i)} &\coloneqq \sum_{v^{(j)} \in \mathcal{V}} \Pi^{\mathrm{PPR}}_{i,j} Q^{\mathrm{ft},(j)}, \label{eq:lopgpn-p}
\end{align}
where $Q^{\mathrm{ft},(j)}$ is the density of the Dirichlet distribution $\mathrm{Dir}(\bm{\alpha}^{\mathrm{ft},(j)})$ and $\Pi^{\mathrm{PPR}}$ as in \cref{eq:gpn-ppr}.
Using such mixtures, \ac{lopgpn} does not assume the irreducibility of conflicts, i.e., unlike \ac{gpn}, \ac{au} is not increased in the presence of conflicts.
This follows trivially from the fact that the \ac{au} of a \ac{lop} distribution is just the weighted linear combination of the \acp{au} of the original distributions, i.e.,
\begin{align}
    \mathrm{AU}(Q^{\mathrm{agg},(i)}) &= 
    \int_{\Delta_K} H(\theta) \, \mathrm{d}Q^{\mathrm{agg},(i)} \nonumber \\
    &= \sum_{v^{(j)}} \Pi^{\mathrm{PPR}}_{i,j} \int_{\Delta_K} H(\theta) \, \mathrm{d}Q^{\mathrm{ft},(j)} \nonumber \\
    &= \sum_{v^{(j)}} \Pi^{\mathrm{PPR}}_{i,j} \mathrm{AU}(Q^{\mathrm{ft},(j)}) . \label{eq:lopgpn-conflict-au}
\end{align}
\Cref{fig:lop-gpn-aggregation} illustrates the implications of this approach for the binary node classification.
Next, we will describe how \ac{lopgpn} is trained and how it can be implemented efficiently.

\subsection{Second-order Loss}\label{sec:lop:loss}

Analogous to \cref{eq:uce}, the loss of \ac{lopgpn} is given by
\begin{align}
    \mathcal{L} \coloneqq \sum_{i=1}^{N} \underbrace{\mathbb{E}_{Q^{\mathrm{agg},(i)}} \left[ \mathrm{CE}(\Theta^{(i)}, y^{(i)}) \right] - H(Q^{\mathrm{agg},(i)})}_{\mathcal{L}^{(i)}} . \label{eq:lopgpn-optim}
\end{align}
Due to the use of Dirichlet mixtures, this loss is not directly minimizable, as there is no closed-form expression for the regularization term $H(Q^{\mathrm{agg},(i)})$.
We therefore use the following bounds on the entropy of a mixture distribution instead (see \citet{melbourne2022}):
\begin{align}
    H(Q^{\mathrm{agg},(i)}) &\geq \smashoperator{\sum_{v^{(j)} \in \mathcal{V}}} \Pi^{\mathrm{PPR}}_{i,j} H(Q^{\mathrm{ft},(j)}), \label{eq:lopgpn-entropy-bound}\\
    H(Q^{\mathrm{agg},(i)}) &\leq H(\mathrm{Cat}(\Pi^{\mathrm{PPR}}_{i})) + \smashoperator{\sum_{v^{(j)} \in \mathcal{V}}} \Pi^{\mathrm{PPR}}_{i,j} H(Q^{\mathrm{ft},(j)}), \nonumber
\end{align}
where $\mathrm{Cat}(\Pi^{\mathrm{PPR}}_{i})$ is the categorical distribution described by the $i^{th}$ row vector of $\Pi^{\mathrm{PPR}}$.
We use the upper bound on $H(Q^{\mathrm{agg},(i)})$ as a surrogate for $\mathrm{EU}_{\mathrm{SO}}$.
Similarly, the lower entropy bound implies the following upper bound on the loss for each vertex $v^{(i)}$:
\begin{align*}
    \mathcal{L}^{(i)} \leq &\smashoperator{\sum_{j=1}^N} \Pi^{\mathrm{PPR}}_{i,j} \left( \mathbb{E}_{Q^{\mathrm{ft},(i)}} \left[ \mathrm{CE}(\Theta^{(i)}, y^{(i)}) \right] - H(Q^{\mathrm{ft},(j)}) \right) 
\end{align*}
This bound on the loss is differentiable and can therefore be minimized using standard gradient-based optimization algorithms.

\subsection{Sparse APPNP}\label{sec:lop:sparsity}

Despite the similarities between \ac{gpn} and \ac{lopgpn}, the computational complexity of the \acs{appnp}-based aggregation step is significantly higher for \ac{lopgpn}.
We can express \cref{eq:gpn-alpha} as a matrix-vector multiplication, i.e.,%
\begin{align}
    \bm{\alpha}^{\mathrm{agg}} &= \Pi^{\mathrm{PPR}} \bm{\alpha}^{\mathrm{ft}} = \hat{A}_{\varepsilon}^L \bm{\alpha}^{\mathrm{ft}},
\end{align}
where $\bm{\alpha}^{\mathrm{agg}}, \bm{\alpha}^{\mathrm{ft}} \in \mathbb{R}_+^{N \times K}$ are pseudo-count matrices and $\hat{A}_{\varepsilon} = \varepsilon \mathbf{I} + (1 - \varepsilon) \mathbf{\hat{A}}$.
Since typically $K \ll N$, it is best to evaluate this product from right to left.
Then, assuming that $A$ is sparse, the complexity of this operation is $\mathcal{O}(L \cdot |\mathcal{E}| \cdot K)$.

\Ac{lopgpn} on the other hand uses the values of $\Pi^{\mathrm{PPR}}$ directly as mixture weights, i.e., the $L$-th power of $\hat{A}_{\varepsilon}$ has to be computed explicitly.
For large graphs, this is computationally infeasible.
To address this issue, a sparse approximation of \ac{appnp} is used to compute $\Pi^{\mathrm{PPR}}$.
Between each of the $L$ sparse matrix multiplications, all probability mass below a certain threshold $\delta$ is moved back to the diagonal of the matrix.
This limits the percentage of non-zero entries to ${(N \delta)}^{-1}$ and makes it possible to apply \ac{lopgpn} even to large graphs.
In the experimental evaluation, sparsification was used for all datasets except CoraML and CiteSeer.

\section{Evaluation}\label{sec:eval}

We evaluate \ac{lopgpn} in two ways:
First, we use \acp{arc} to compare the quality of different predictive uncertainty measures of \ac{lopgpn} to those of \ac{gpn} on a set of standard node classification benchmarks.
Second, we compare the \ac{ood} detection performance of our model against a set of node classification models.

\subsection{Experimental Setup}\label{sec:eval:setup}

Due to the similarity between \ac{lopgpn} and \ac{gpn}, we base our experiments on those of \citet{stadler2021}, i.e., we use the same dataset splits and hyperparameters and build upon their reference implementation.\footnote{Implementation available at \url{https://github.com/Cortys/gpn-extensions}}

\paragraph{Datasets}
We use the following node classification benchmarks:
Three citation network datasets, namely, \textbf{CoraML}, \textbf{CiteSeer} and \textbf{PubMed}~\citep{mccallum2000,giles1998,getoor2005,sen2008,namata2012},
two co-purchase datasets, namely, \textbf{Amazon Photos} and \textbf{Amazon Computers}~\citep{mcauley2015}
and the large-scale \textbf{OGBN Arxiv} dataset with about $170 \mathrm{k}$ nodes and over $2.3$ million edges~\citep{hu2020}.
Since OGBN Arxiv is presplit into train, validation and test sets, we use the provided splits.
The results of the other datasets are obtained by averaging over 10 dataset splits with train/val/test sizes of 5\%/15\%/80\%.
Note that all six datasets represent either citation networks or co-purchase networks; the assumptions of unbounded growth of neighborhood sizes and network homophily (see \cref{sec:uq:critique}) are therefore reasonable and the use of \ac{lopgpn} well-motivated.

\paragraph{Models}
We compare \textbf{\ac{lopgpn}} against the following baseline models:
Two variants of \textbf{\ac{gpn}}~\citep{stadler2021}, \textbf{\ac{appnp}}~\citep{gasteiger2018}, \textbf{Matern-GGP}~\citep{borovitskiy2021} and \textbf{GKDE}~\citep{zhao2020}.

As described in \cref{sec:uq:gpn}, \textbf{\ac{appnp}} directly disperses class probabilities, i.e., it is a first-order method that cannot (meaningfully) distinguish between \ac{au} and \ac{eu}; the entropy of its first-order predictions is therefore used as an estimate of \acl{tu}.

\textbf{Matern-GGP}~\citep{bojchevski2018} is a Gaussian process model using the so-called \emph{graph Matérn kernel}.
For each vertex, it predicts a posterior second-order multivariate Gaussian $\Theta \sim \mathcal{N}(\bar{\theta}, \Sigma)$ with $\bar{\theta} \in \Delta_K$ and $\Sigma$ a diagonal covariance matrix.
This implies that some probability mass will lie outside of $\Delta_K$; thus, the additive entropy-based uncertainty decomposition described in \cref{sec:um:entropy} is not well-defined for this distribution.
We therefore do not use the additive entropy decomposition for Matern-GGP and instead use $H(\bar{\theta})$ as a proxy for \ac{au} and the trace $\mathrm{Tr}(\Sigma)$ as a proxy for \ac{eu}.
Note that this definition of \ac{au} coincides with the definition of \ac{tu} in the additive decomposition (\cref{eq:tu}).
Second, note that $\mathrm{Tr}(\Sigma)$ is a monotonic transformation of the differential entropy of the second-order Gaussian $\mathcal{N}(\bar{\theta}, \Sigma)$ and thereby closely related to $\mathrm{EU}_{\mathrm{SO}}$ (\cref{eq:eu-so-entropy}).\footnote{%
The differential entropy of the multinomial Gaussian $\mathcal{N}(\bar{\theta}, \Sigma)$ is $\frac{1}{2}\ln\left((2 \pi e)^k \det \Sigma \right) = \frac{1}{2}\mathrm{Tr}(\ln \Sigma) + c \geq \frac{1}{2}\ln\mathrm{Tr}(\Sigma) + c$ with the constant $c = \frac{k}{2}\ln(2 \pi e)$.
}
Last, we note that the Matern-GGP model was not applied to the large-scale OGBN Arxiv dataset due to memory constraints.

The \emph{Graph-based Kernel Dirichlet distribution Estimation} \textbf{GKDE} model~\citep{zhao2020}, is a parameter-free method that estimates a Dirichlet distribution for each node based on the features of its neighbors.

The two evaluated variants of \textbf{\ac{gpn}} are \emph{\ac{gpn} (rw)} and \emph{\ac{gpn} (sym)}.
\ac{gpn} (rw) uses \acs{appnp} with random-walk normalization, i.e., with $\hat{\mathbf{A}} = \mathbf{A} \mathbf{D}^{-1}$.
\ac{gpn} (sym) uses symmetric normalization, i.e., with $\hat{\mathbf{A}} = \mathbf{D}^{-1/2} \mathbf{A} \mathbf{D}^{-1/2}$ (see \citet{kipf2017}).
We evaluate both types of normalization because \citet{stadler2021} used symmetric normalization in their experiments, while \ac{lopgpn} requires random-walk normalization to ensure that valid mixture densities are produced.
To show that any observed differences between \ac{lopgpn} and \ac{gpn} are due to the use of \ac{lop} and not due to the use of random-walk normalization, we compare \ac{lopgpn} against both variants.

\subsection{Accuracy-Rejection Curves}\label{sec:eval:arc}

\newcommand*{\accrejPlot}[4]{\begin{tikzpicture}[inner frame sep=0]
    \small
    \begin{axis}[
        width=0.2\linewidth,
        height=2.9cm,
        xlabel={\ifthenelse{\equal{#1}{Arxiv}}{#4}{}},
        ylabel={\ifthenelse{\equal{#2}{sample_total_entropy}}{\scriptsize\ifthenelse{\equal{#1}{Computers}\or\equal{#1}{Photos}}{A.#1}{\ifthenelse{\equal{#1}{Arxiv}}{OGBN #1}{#1}}}{}},
        grid=major,
        ymin=#3, ymax=1,
        xmin=0, xmax=0.99,
        label style={font=\footnotesize},
        tick label style={font=\tiny},
        y tick label style={/pgf/number format/.cd, fixed, fixed zerofill, precision=2, /tikz/.cd},
        legend pos=outer north east,
        legend style={font=\tiny, draw=none},
        legend image post style={line width=2pt},
    ]

    \addplot [color=t_green, line width=1pt] table [x=p, y=gpnMean, col sep=comma] {tables/acc_rej_#2_#1.csv};
    \addplot [color=t_blue, line width=1pt] table [x=p, y=gpnRWMean, col sep=comma] {tables/acc_rej_#2_#1.csv};
    \addplot [color=t_red, line width=1pt] table [x=p, y=gpnLOPMean, col sep=comma] {tables/acc_rej_#2_#1.csv};

    \addSE{tables/acc_rej_#2_#1.csv}{gpn}{t_green}
    \addSE{tables/acc_rej_#2_#1.csv}{gpnRW}{t_blue}
    \addSE{tables/acc_rej_#2_#1.csv}{gpnLOP}{t_red}

    \ifthenelse{\equal{#2}{sample_epistemic_entropy}}{%
        \legend{GPN (sym), GPN (rw), LOP-GPN}%
    }{}
    \end{axis}
\end{tikzpicture}}
\newcommand*{\accrejPlots}[2]{
    \hfill\accrejPlot{#1}{sample_total_entropy}{#2}{$\mathrm{TU}$}%
    \hfill\accrejPlot{#1}{sample_aleatoric_entropy}{#2}{$\mathrm{AU}$}%
    \hfill\accrejPlot{#1}{sample_epistemic_entropy_diff}{#2}{$\mathrm{EU}$}%
    \hfill\accrejPlot{#1}{sample_epistemic}{#2}{$\mathrm{EU}_{\mathrm{PC}}$}%
    \hfill\accrejPlot{#1}{sample_epistemic_entropy}{#2}{$\mathrm{EU}_{\mathrm{SO}}$}%
}
\begin{figure*}[ht!]
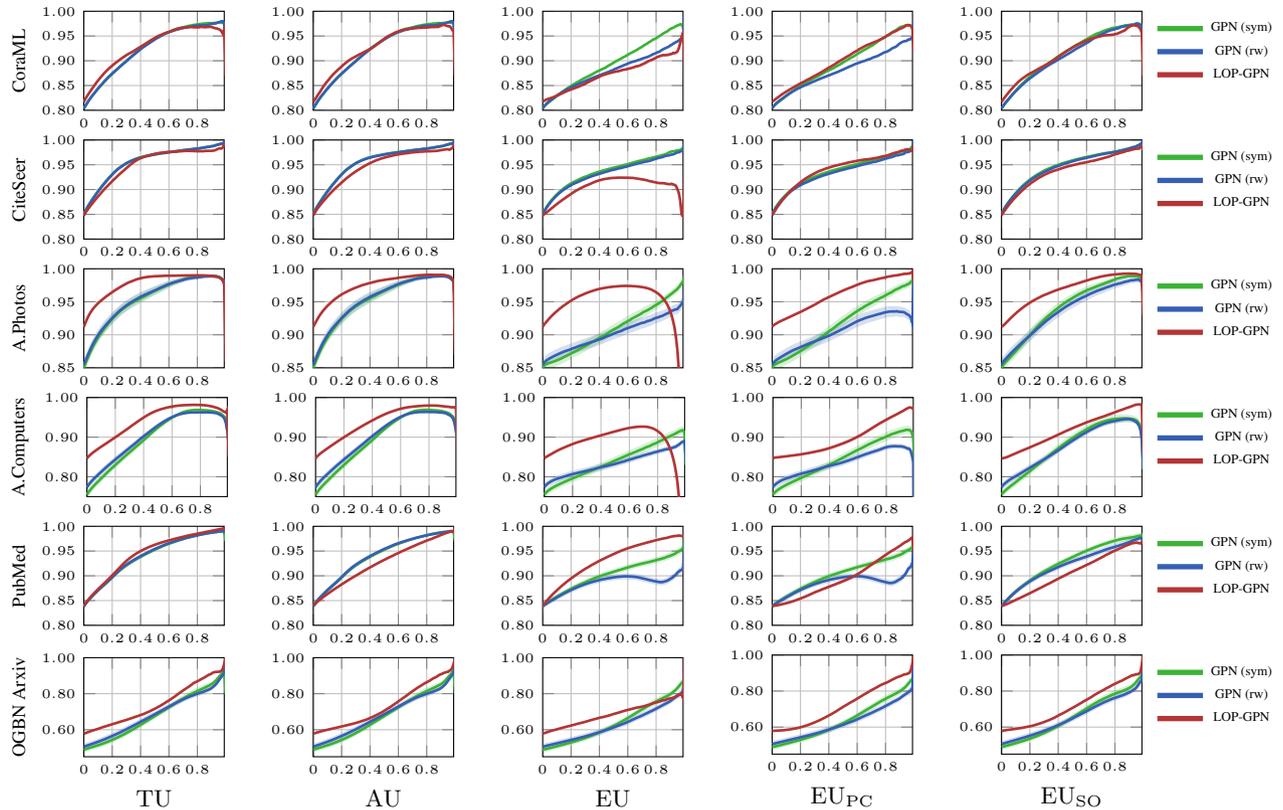

    \centering
    \accrejPlots{CoraML}{0.8}\\[-10pt]%
    \accrejPlots{CiteSeer}{0.8}\\[-10pt]%
    \accrejPlots{Photos}{0.85}\\[-10pt]%
    \accrejPlots{Computers}{0.75}\\[-10pt]%
    \accrejPlots{PubMed}{0.8}\\[-10pt]%
    \accrejPlots{Arxiv}{0.45}%
    \caption{%
        \Acl*{arc} for different uncertainty measures.
        The x-axis represents the fraction of rejected test instances; the y-axis represents the test accuracy for a given rejection rate.
        The (small) shaded areas behind the curves represent the estimate's standard error.
    }
    \label{fig:accrej}
\end{figure*}
\Acfp{arc} are produced by discarding the predictions for instances where the predictor exhibits the highest uncertainty, and then calculating the accuracy for the remaining subset.
For an uncertainty measure which captures predictive uncertainty well, the accuracy should monotonically increase with the rejection rate.
We use the following uncertainty measures to produce \acp{arc}:
Entropy-based \ac{tu}, \ac{au} and \ac{eu} (\cref{eq:tu,eq:au,eq:eu}),
second-order \acl{eu} (\cref{eq:eu-so-entropy}),
and pseudo-count-based \acl{eu}.
\Cref{fig:accrej} show the \acp{arc} for the different uncertainty measures for the six evaluated datasets.
First, note that \ac{lopgpn} consistently outperforms or at least matches \ac{gpn} for almost all rejection rates and all datasets.
This indicates that dropping the assumption of the \emph{irreducibility of conflicts} made by \ac{gpn} is beneficial in practice.
The only exception to this are the curves for the mutual information-based \ac{eu} on the CiteSeer, Amazon Photos and Amazon Computers datasets; here, the accuracies of \ac{lopgpn} drop-off at high rejection rates.
As previously illustrated in \cref{fig:entropy-uncertainty}, the mutual information-based \ac{eu} can be smaller for the uninformed uniform second-order distribution than for a (more-informed) bimodal Dirichlet mixture distribution.
Analogous to that example, we hypothesize that some of the uninformed mixture distributions produced by \ac{lopgpn} are incorrectly assigned low \ac{eu} values, leading to the observed drop-off in accuracy for the nodes with the lowest \ac{eu} estimates.
The (arguably) more reasonable pseudo-count- and differential entropy-based \ac{eu} measures do not exhibit this behavior.
There, the accuracies go up for increasing rejection rates; this is a sign that these measures capture predictive uncertainty in a meaningful way.

\subsection{Out-of-Distribution Detection}\label{sec:eval:ood}

\begin{table*}[ht!]
    \caption{\Acs*{ood} detection performance of \acs*{ood} vs \acs*{id} vertices and \acs*{id} accuracies for three \acs*{ood} scenarios.}
    \label{tbl:ood}
    \centering
    {\tiny%
    \begin{filecontents*}{tables/id_ood.csv}
id,dataset,model,acc,accSE,accBest,oodLocIdAcc,oodLocIdAccSE,oodLocIdAccBest,oodLocOodAcc,oodLocOodAccSE,oodLocOodAccBest,oodLocOodTotal,oodLocOodTotalSE,oodLocOodTotalBest,oodLocOodTotalEntropy,oodLocOodTotalEntropySE,oodLocOodTotalEntropyBest,oodLocOodAleatoric,oodLocOodAleatoricSE,oodLocOodAleatoricBest,oodLocOodAleatoricEntropy,oodLocOodAleatoricEntropySE,oodLocOodAleatoricEntropyBest,oodLocOodEpistemic,oodLocOodEpistemicSE,oodLocOodEpistemicBest,oodLocOodEpistemicEntropy,oodLocOodEpistemicEntropySE,oodLocOodEpistemicEntropyBest,oodLocOodEpistemicEntropyDiff,oodLocOodEpistemicEntropyDiffSE,oodLocOodEpistemicEntropyDiffBest,oodLocTotalEntropyChange,oodLocAleatoricEntropyChange,oodLocEpistemicEntropyChange,oodLocEpistemicEntropyDiffChange,oodNormalIdAcc,oodNormalIdAccSE,oodNormalIdAccBest,oodNormalOodAcc,oodNormalOodAccSE,oodNormalOodAccBest,oodNormalOodTotal,oodNormalOodTotalSE,oodNormalOodTotalBest,oodNormalOodTotalEntropy,oodNormalOodTotalEntropySE,oodNormalOodTotalEntropyBest,oodNormalOodAleatoric,oodNormalOodAleatoricSE,oodNormalOodAleatoricBest,oodNormalOodAleatoricEntropy,oodNormalOodAleatoricEntropySE,oodNormalOodAleatoricEntropyBest,oodNormalOodEpistemic,oodNormalOodEpistemicSE,oodNormalOodEpistemicBest,oodNormalOodEpistemicEntropy,oodNormalOodEpistemicEntropySE,oodNormalOodEpistemicEntropyBest,oodNormalOodEpistemicEntropyDiff,oodNormalOodEpistemicEntropyDiffSE,oodNormalOodEpistemicEntropyDiffBest,oodNormalTotalEntropyChange,oodNormalAleatoricEntropyChange,oodNormalEpistemicEntropyChange,oodNormalEpistemicEntropyDiffChange,oodBerIdAcc,oodBerIdAccSE,oodBerIdAccBest,oodBerOodAcc,oodBerOodAccSE,oodBerOodAccBest,oodBerOodTotal,oodBerOodTotalSE,oodBerOodTotalBest,oodBerOodTotalEntropy,oodBerOodTotalEntropySE,oodBerOodTotalEntropyBest,oodBerOodAleatoric,oodBerOodAleatoricSE,oodBerOodAleatoricBest,oodBerOodAleatoricEntropy,oodBerOodAleatoricEntropySE,oodBerOodAleatoricEntropyBest,oodBerOodEpistemic,oodBerOodEpistemicSE,oodBerOodEpistemicBest,oodBerOodEpistemicEntropy,oodBerOodEpistemicEntropySE,oodBerOodEpistemicEntropyBest,oodBerOodEpistemicEntropyDiff,oodBerOodEpistemicEntropyDiffSE,oodBerOodEpistemicEntropyDiffBest,oodBerTotalEntropyChange,oodBerAleatoricEntropyChange,oodBerEpistemicEntropyChange,oodBerEpistemicEntropyDiffChange
0,CoraML,APPNP,84.35,0.00,1,90.44,0.00,1,0.00,0.00,1,86.40,0.00,1,87.45,0.00,1,86.40,0.00,1,,0.00,0,,0.00,0,,0.00,0,,0.00,0,+125.64,+125.64,,,43.62,0.00,0,18.69,0.00,0,14.48,0.00,0,13.26,0.00,0,14.48,0.00,0,,0.00,0,,0.00,0,,0.00,0,,0.00,0,-84.97,-84.97,,,84.14,0.00,1,80.60,0.00,1,63.85,0.00,1,65.02,0.00,1,63.85,0.00,1,,0.00,0,,0.00,0,,0.00,0,,0.00,0,,,,
1,CoraML,Matern-GGP,77.17,0.00,0,85.49,0.00,0,0.00,0.00,1,82.02,0.00,0,82.31,0.00,0,82.02,0.00,0,82.31,0.00,0,,0.00,0,82.18,0.00,0,,0.00,0,+122.25,+122.25,,,77.15,0.00,0,77.36,0.00,0,49.88,0.00,0,49.71,0.00,0,49.88,0.00,0,49.71,0.00,0,,0.00,0,49.81,0.00,0,,0.00,0,-1.287,-1.287,,,77.15,0.00,0,77.36,0.00,0,49.88,0.00,0,49.71,0.00,0,49.88,0.00,0,49.71,0.00,0,,0.00,0,49.81,0.00,0,,0.00,0,,,,
2,CoraML,GKDE,71.73,0.00,0,83.01,0.00,0,0.00,0.00,1,77.65,0.00,0,77.21,0.00,0,77.65,0.00,0,35.35,0.00,0,74.00,0.00,0,76.46,0.00,0,69.16,0.00,0,+0.400,-0.401,-198.1,+3.426,71.96,0.00,0,69.69,0.00,0,48.83,0.00,0,48.71,0.00,0,48.83,0.00,0,50.76,0.00,0,48.45,0.00,0,48.68,0.00,0,49.19,0.00,0,-0.003,+0.017,-200.0,-0.100,71.96,0.00,0,69.69,0.00,0,48.83,0.00,0,48.71,0.00,0,48.83,0.00,0,50.76,0.00,0,48.45,0.00,0,48.68,0.00,0,49.19,0.00,0,,,-200.0,-0.100
3,CoraML,GPN (sym),80.34,0.00,0,89.36,0.00,0,0.00,0.00,1,84.29,0.00,0,85.51,0.00,0,84.29,0.00,0,85.49,0.00,0,87.11,0.00,1,89.15,0.00,1,86.23,0.00,1,+104.49,+104.42,-174.9,+229.74,17.86,0.00,0,18.19,0.00,0,89.14,0.00,1,96.59,0.00,1,89.14,0.00,1,96.80,0.00,1,70.62,0.00,1,75.66,0.00,0,70.89,0.00,0,+2.038,+2.018,-196.9,+24.074,80.36,0.00,0,78.41,0.00,0,54.17,0.00,0,54.44,0.00,0,54.17,0.00,0,54.44,0.00,0,54.17,0.00,0,87.19,0.00,0,51.56,0.00,0,,,-176.3,+15.300
4,CoraML,GPN (rw),80.36,0.00,0,89.30,0.00,0,0.00,0.00,1,83.96,0.00,0,85.19,0.00,0,83.96,0.00,0,85.17,0.00,0,83.11,0.00,0,87.60,0.00,0,82.52,0.00,0,+104.22,+104.15,-174.7,+225.28,17.86,0.00,0,18.19,0.00,0,87.34,0.00,0,92.65,0.00,0,87.34,0.00,0,93.22,0.00,0,63.21,0.00,0,66.91,0.00,0,63.40,0.00,0,+2.431,+2.401,-196.5,+32.053,80.51,0.00,0,78.12,0.00,0,54.67,0.00,0,55.09,0.00,0,54.67,0.00,0,55.09,0.00,0,54.32,0.00,0,87.67,0.00,1,51.88,0.00,0,,,-176.1,+16.544
5,CoraML,LOP-GPN,81.69,0.00,0,89.34,0.00,0,0.00,0.00,1,84.45,0.00,0,85.67,0.00,0,84.45,0.00,0,88.51,0.00,1,84.72,0.00,0,79.26,0.00,0,45.18,0.00,0,+56.987,+81.573,-174.8,-5.327,81.74,0.00,1,79.92,0.00,1,65.19,0.00,0,69.50,0.00,0,65.19,0.00,0,61.06,0.00,0,60.15,0.00,0,83.69,0.00,1,82.08,0.00,1,+20.342,+15.113,-180.7,+33.451,81.29,0.00,0,76.82,0.00,0,62.04,0.00,0,62.20,0.00,0,62.04,0.00,0,59.69,0.00,1,54.40,0.00,1,56.74,0.00,0,64.31,0.00,1,,,-195.8,+16.870
6,CiteSeer,APPNP,86.04,0.00,1,87.75,0.00,1,0.00,0.00,1,85.33,0.00,1,85.80,0.00,1,85.33,0.00,1,,0.00,0,,0.00,0,,0.00,0,,0.00,0,+159.54,+159.54,,,72.72,0.00,0,25.05,0.00,0,26.62,0.00,0,23.76,0.00,0,26.62,0.00,0,,0.00,0,,0.00,0,,0.00,0,,0.00,0,-62.75,-62.75,,,85.65,0.00,1,83.64,0.00,1,68.65,0.00,0,70.25,0.00,0,68.65,0.00,0,,0.00,0,,0.00,0,,0.00,0,,0.00,0,,,,
7,CiteSeer,Matern-GGP,50.15,0.00,0,49.05,0.00,0,0.00,0.00,1,78.70,0.00,0,78.56,0.00,0,78.70,0.00,0,78.56,0.00,0,,0.00,0,78.59,0.00,0,,0.00,0,+152.87,+152.87,,,50.02,0.00,0,51.28,0.00,0,50.52,0.00,0,50.56,0.00,0,50.52,0.00,0,50.56,0.00,0,,0.00,0,50.55,0.00,0,,0.00,0,+1.041,+1.041,,,50.02,0.00,0,51.28,0.00,0,50.52,0.00,0,50.56,0.00,0,50.52,0.00,0,50.56,0.00,0,,0.00,0,50.55,0.00,0,,0.00,0,,,,
8,CiteSeer,GKDE,65.85,0.00,0,73.73,0.00,0,0.00,0.00,1,81.33,0.00,0,77.89,0.00,0,81.33,0.00,0,37.03,0.00,0,81.30,0.00,1,79.44,0.00,0,65.08,0.00,0,+0.131,-0.291,-199.3,+1.673,65.79,0.00,0,66.45,0.00,0,50.18,0.00,0,50.14,0.00,0,50.18,0.00,0,51.33,0.00,0,50.17,0.00,0,50.04,0.00,0,48.80,0.00,0,+0.002,+0.019,-199.9,-0.070,65.79,0.00,0,66.45,0.00,0,50.18,0.00,0,50.14,0.00,0,50.18,0.00,0,51.33,0.00,0,50.17,0.00,0,50.04,0.00,0,48.80,0.00,0,,,-199.9,-0.070
9,CiteSeer,GPN (sym),85.13,0.00,0,87.23,0.00,0,0.00,0.00,1,80.87,0.00,0,81.53,0.00,0,80.87,0.00,0,81.52,0.00,1,76.76,0.00,0,80.62,0.00,1,76.78,0.00,1,+98.306,+98.324,-181.8,+84.116,17.45,0.00,0,17.30,0.00,0,86.96,0.00,1,93.00,0.00,1,86.96,0.00,1,93.22,0.00,1,79.91,0.00,1,86.34,0.00,0,80.32,0.00,0,+5.183,+5.158,-195.1,+8.989,84.82,0.00,0,82.24,0.00,0,61.82,0.00,0,62.58,0.00,0,61.82,0.00,0,62.58,0.00,0,60.16,0.00,1,91.73,0.00,1,53.65,0.00,0,,,-171.7,+32.222
10,CiteSeer,GPN (rw),85.03,0.00,0,87.12,0.00,0,0.00,0.00,1,80.61,0.00,0,81.23,0.00,0,80.61,0.00,0,81.23,0.00,0,74.26,0.00,0,79.61,0.00,0,74.29,0.00,0,+98.668,+98.689,-181.6,+82.022,17.45,0.00,0,17.30,0.00,0,85.80,0.00,0,91.28,0.00,0,85.80,0.00,0,91.68,0.00,0,72.00,0.00,0,78.25,0.00,0,72.38,0.00,0,+5.527,+5.496,-194.9,+10.098,84.78,0.00,0,81.97,0.00,0,62.39,0.00,0,63.16,0.00,0,62.39,0.00,0,63.16,0.00,0,60.15,0.00,0,91.59,0.00,0,53.78,0.00,0,,,-171.4,+34.481
11,CiteSeer,LOP-GPN,84.74,0.00,0,87.16,0.00,0,0.00,0.00,1,81.66,0.00,0,82.19,0.00,0,81.66,0.00,0,80.57,0.00,0,75.48,0.00,0,73.12,0.00,0,73.09,0.00,0,+80.393,+77.788,-185.4,+84.756,84.41,0.00,1,82.74,0.00,1,75.82,0.00,0,81.65,0.00,0,75.82,0.00,0,76.92,0.00,0,59.20,0.00,0,86.81,0.00,1,81.26,0.00,1,+55.932,+54.267,-172.5,+58.547,84.31,0.00,0,81.37,0.00,0,75.14,0.00,1,78.74,0.00,1,75.14,0.00,1,78.43,0.00,1,58.12,0.00,0,68.98,0.00,0,68.54,0.00,1,,,-189.9,+33.167
12,Amazon\\Photos,APPNP,91.68,0.00,1,94.99,0.00,1,0.00,0.00,1,75.58,0.00,0,75.12,0.00,0,75.58,0.00,0,,0.00,0,,0.00,0,,0.00,0,,0.00,0,+79.465,+79.465,,,40.44,0.00,0,19.84,0.00,0,14.42,0.00,0,13.17,0.00,0,14.42,0.00,0,,0.00,0,,0.00,0,,0.00,0,,0.00,0,-85.22,-85.22,,,91.83,0.00,1,86.76,0.00,1,63.58,0.00,0,63.55,0.00,0,63.58,0.00,0,,0.00,0,,0.00,0,,0.00,0,,0.00,0,,,,
13,Amazon\\Photos,Matern-GGP,85.99,0.00,0,88.57,0.00,0,0.00,0.00,1,82.81,0.00,0,82.12,0.00,0,82.81,0.00,0,82.12,0.00,0,,0.00,0,82.68,0.00,0,,0.00,0,+184.66,+184.66,,,86.04,0.00,0,85.49,0.00,0,49.63,0.00,0,49.62,0.00,0,49.63,0.00,0,49.62,0.00,0,,0.00,0,49.62,0.00,0,,0.00,0,-5.903,-5.903,,,86.04,0.00,0,85.49,0.00,0,49.63,0.00,0,49.62,0.00,0,49.63,0.00,0,49.62,0.00,0,,0.00,0,49.62,0.00,0,,0.00,0,,,,
14,Amazon\\Photos,GKDE,76.16,0.00,0,85.45,0.00,0,0.00,0.00,1,75.92,0.00,0,70.20,0.00,0,75.92,0.00,0,55.45,0.00,0,60.80,0.00,0,66.83,0.00,0,61.23,0.00,0,+1.361,+0.435,-196.3,+5.897,76.19,0.00,0,75.87,0.00,0,49.21,0.00,0,49.07,0.00,0,49.21,0.00,0,50.74,0.00,0,48.87,0.00,0,48.95,0.00,0,48.76,0.00,0,-0.064,+0.047,-200.2,-0.729,76.19,0.00,0,75.87,0.00,0,49.21,0.00,0,49.07,0.00,0,49.21,0.00,0,50.74,0.00,0,48.87,0.00,0,48.95,0.00,0,48.76,0.00,0,,,-200.2,-0.729
15,Amazon\\Photos,GPN (sym),85.02,0.00,0,91.49,0.00,0,0.00,0.00,1,75.39,0.00,0,76.29,0.00,0,75.39,0.00,0,76.29,0.00,0,87.50,0.00,1,86.05,0.00,1,86.54,0.00,1,+57.315,+57.311,-184.5,+196.11,12.63,0.00,0,12.83,0.00,0,95.40,0.00,1,89.43,0.00,0,95.40,0.00,1,91.07,0.00,1,59.86,0.00,0,63.56,0.00,0,60.06,0.00,0,+1.613,+1.579,-197.4,+17.020,84.79,0.00,0,81.95,0.00,0,54.36,0.00,0,54.55,0.00,0,54.36,0.00,0,54.55,0.00,0,54.29,0.00,1,80.17,0.00,1,49.87,0.00,0,,,-179.2,+6.485
16,Amazon\\Photos,GPN (rw),85.45,0.00,0,91.76,0.00,0,0.00,0.00,1,75.49,0.00,0,76.22,0.00,0,75.49,0.00,0,76.22,0.00,0,78.71,0.00,0,81.72,0.00,0,77.93,0.00,0,+58.174,+58.170,-185.3,+173.91,12.63,0.00,0,12.87,0.00,0,90.30,0.00,0,79.70,0.00,0,90.30,0.00,0,81.36,0.00,0,55.99,0.00,0,58.16,0.00,0,56.12,0.00,0,+2.087,+2.023,-197.0,+27.141,85.35,0.00,0,82.26,0.00,0,55.25,0.00,0,55.38,0.00,0,55.25,0.00,0,55.38,0.00,0,53.33,0.00,0,74.95,0.00,0,49.92,0.00,0,,,-182.7,+6.096
17,Amazon\\Photos,LOP-GPN,91.19,0.00,0,94.00,0.00,0,0.00,0.00,1,83.86,0.00,1,86.50,0.00,1,83.86,0.00,1,83.88,0.00,1,76.63,0.00,0,83.87,0.00,0,80.02,0.00,0,+113.05,+111.07,-184.7,+116.42,91.18,0.00,1,86.89,0.00,1,84.54,0.00,0,91.76,0.00,1,84.54,0.00,0,87.29,0.00,0,61.15,0.00,1,95.74,0.00,1,89.36,0.00,1,+93.379,+98.786,-165.4,+85.072,91.10,0.00,0,77.69,0.00,0,79.78,0.00,1,79.21,0.00,1,79.78,0.00,1,70.67,0.00,1,53.39,0.00,0,61.00,0.00,0,78.59,0.00,1,,,-195.6,+69.456
18,Amazon\\Computers,APPNP,80.92,0.00,0,87.99,0.00,0,0.00,0.00,1,80.31,0.00,0,79.32,0.00,0,80.31,0.00,0,,0.00,0,,0.00,0,,0.00,0,,0.00,0,+76.515,+76.515,,,42.81,0.00,0,21.64,0.00,0,16.93,0.00,0,15.58,0.00,0,16.93,0.00,0,,0.00,0,,0.00,0,,0.00,0,,0.00,0,-82.43,-82.43,,,81.11,0.00,0,75.97,0.00,0,64.20,0.00,0,64.62,0.00,0,64.20,0.00,0,,0.00,0,,0.00,0,,0.00,0,,0.00,0,,,,
19,Amazon\\Computers,Matern-GGP,81.04,0.00,0,86.74,0.00,0,0.00,0.00,1,81.83,0.00,0,82.20,0.00,0,81.83,0.00,0,82.20,0.00,0,,0.00,0,81.94,0.00,0,,0.00,0,+185.59,+185.59,,,81.00,0.00,0,81.37,0.00,0,50.00,0.00,0,50.02,0.00,0,50.00,0.00,0,50.02,0.00,0,,0.00,0,50.00,0.00,0,,0.00,0,+2.427,+2.427,,,81.00,0.00,0,81.37,0.00,1,50.00,0.00,0,50.02,0.00,0,50.00,0.00,0,50.02,0.00,0,,0.00,0,50.00,0.00,0,,0.00,0,,,,
20,Amazon\\Computers,GKDE,64.04,0.00,0,71.26,0.00,0,0.00,0.00,1,75.20,0.00,0,76.38,0.00,0,75.20,0.00,0,70.52,0.00,0,74.37,0.00,0,76.20,0.00,0,74.46,0.00,0,+9.293,+5.394,-180.5,+35.198,64.01,0.00,0,64.34,0.00,0,49.88,0.00,0,49.92,0.00,0,49.88,0.00,0,49.81,0.00,0,50.09,0.00,0,49.99,0.00,0,50.03,0.00,0,-0.027,-0.034,-200.0,+0.026,64.01,0.00,0,64.34,0.00,0,49.88,0.00,0,49.92,0.00,0,49.88,0.00,0,49.81,0.00,0,50.09,0.00,0,49.99,0.00,0,50.03,0.00,0,,,-200.0,+0.026
21,Amazon\\Computers,GPN (sym),75.53,0.00,0,82.17,0.00,0,0.00,0.00,1,77.17,0.00,0,79.17,0.00,0,77.17,0.00,0,79.17,0.00,0,81.01,0.00,1,83.77,0.00,0,76.65,0.00,1,+44.150,+44.151,-189.0,+15.961,16.39,0.00,0,17.03,0.00,0,83.24,0.00,0,88.69,0.00,0,83.24,0.00,0,89.62,0.00,0,62.12,0.00,0,65.90,0.00,0,62.32,0.00,0,+2.973,+2.919,-197.0,+19.750,75.71,0.00,0,74.51,0.00,0,54.28,0.00,0,54.67,0.00,0,54.28,0.00,0,54.67,0.00,0,56.46,0.00,1,85.57,0.00,0,51.14,0.00,0,,,-174.9,+9.359
22,Amazon\\Computers,GPN (rw),77.38,0.00,0,84.45,0.00,0,0.00,0.00,1,77.72,0.00,0,79.45,0.00,0,77.72,0.00,0,79.45,0.00,0,76.01,0.00,0,81.79,0.00,0,72.80,0.00,0,+49.176,+49.176,-188.2,+68.606,16.39,0.00,0,17.24,0.00,0,79.06,0.00,0,79.77,0.00,0,79.06,0.00,0,80.82,0.00,0,58.25,0.00,0,60.57,0.00,0,58.38,0.00,0,+3.851,+3.752,-196.3,+30.708,76.47,0.00,0,75.02,0.00,0,54.92,0.00,0,55.44,0.00,0,54.92,0.00,0,55.45,0.00,0,55.25,0.00,0,86.11,0.00,1,51.15,0.00,0,,,-174.2,+10.444
23,Amazon\\Computers,LOP-GPN,84.62,0.00,1,90.28,0.00,1,0.00,0.00,1,83.24,0.00,1,85.00,0.00,1,83.24,0.00,1,88.08,0.00,1,77.98,0.00,0,83.80,0.00,1,68.40,0.00,0,+85.739,+105.73,-189.0,+48.044,84.49,0.00,1,81.42,0.00,1,86.47,0.00,1,94.76,0.00,1,86.47,0.00,1,90.93,0.00,1,65.30,0.00,1,98.15,0.00,1,88.33,0.00,1,+84.477,+87.874,-162.6,+78.112,84.35,0.00,1,67.95,0.00,0,79.62,0.00,1,79.78,0.00,1,79.62,0.00,1,72.26,0.00,1,51.59,0.00,0,61.91,0.00,0,75.86,0.00,1,,,-197.2,+59.183
24,PubMed,APPNP,86.63,0.00,1,94.59,0.00,1,0.00,0.00,1,69.38,0.00,0,69.38,0.00,0,69.38,0.00,0,,0.00,0,,0.00,0,,0.00,0,,0.00,0,+84.405,+84.405,,,59.59,0.00,0,37.81,0.00,0,12.47,0.00,0,10.95,0.00,0,12.47,0.00,0,,0.00,0,,0.00,0,,0.00,0,,0.00,0,-86.78,-86.78,,,86.73,0.00,1,82.75,0.00,1,64.97,0.00,0,66.90,0.00,0,64.97,0.00,0,,0.00,0,,0.00,0,,0.00,0,,0.00,0,,,,
25,PubMed,Matern-GGP,78.65,0.00,0,90.05,0.00,0,0.00,0.00,1,63.77,0.00,0,63.72,0.00,0,63.77,0.00,0,63.72,0.00,0,,0.00,0,63.72,0.00,0,,0.00,0,+46.754,+46.754,,,78.62,0.00,0,78.92,0.00,0,50.14,0.00,0,50.14,0.00,0,50.14,0.00,0,50.14,0.00,0,,0.00,0,50.13,0.00,0,,0.00,0,+0.515,+0.515,,,78.62,0.00,0,78.92,0.00,0,50.14,0.00,0,50.14,0.00,0,50.14,0.00,0,50.14,0.00,0,,0.00,0,50.13,0.00,0,,0.00,0,,,,
26,PubMed,GKDE,76.99,0.00,0,87.93,0.00,0,0.00,0.00,1,71.50,0.00,1,71.51,0.00,1,71.50,0.00,1,39.69,0.00,0,69.74,0.00,0,71.00,0.00,0,65.30,0.00,0,+0.774,-0.536,-120.3,+4.457,76.96,0.00,0,77.23,0.00,0,49.94,0.00,0,49.94,0.00,0,49.94,0.00,0,50.16,0.00,0,49.87,0.00,0,49.93,0.00,0,49.79,0.00,0,-0.014,-0.002,-200.0,-0.057,76.96,0.00,0,77.23,0.00,0,49.94,0.00,0,49.94,0.00,0,49.94,0.00,0,50.16,0.00,0,49.87,0.00,0,49.93,0.00,0,49.79,0.00,0,,,-200.0,-0.057
27,PubMed,GPN (sym),83.85,0.00,0,93.76,0.00,0,0.00,0.00,1,69.87,0.00,0,69.87,0.00,0,69.87,0.00,0,69.85,0.00,1,72.93,0.00,1,73.62,0.00,1,72.95,0.00,1,+57.817,+57.809,-185.8,+64.653,30.29,0.00,0,30.44,0.00,0,93.44,0.00,1,95.74,0.00,1,93.44,0.00,1,96.41,0.00,1,70.29,0.00,1,76.44,0.00,0,70.52,0.00,0,+4.252,+4.236,-196.3,+6.699,83.89,0.00,0,81.16,0.00,0,58.10,0.00,0,58.49,0.00,0,58.10,0.00,0,58.49,0.00,0,60.66,0.00,1,81.42,0.00,1,51.12,0.00,0,,,-175.0,+15.346
28,PubMed,GPN (rw),83.86,0.00,0,94.06,0.00,0,0.00,0.00,1,69.52,0.00,0,69.52,0.00,0,69.52,0.00,0,69.50,0.00,0,67.66,0.00,0,72.20,0.00,0,67.69,0.00,0,+57.036,+57.021,-185.8,+68.431,30.29,0.00,0,30.44,0.00,0,89.58,0.00,0,91.19,0.00,0,89.58,0.00,0,92.56,0.00,0,65.92,0.00,0,70.80,0.00,0,66.13,0.00,0,+5.191,+5.162,-195.6,+9.129,83.94,0.00,0,80.91,0.00,0,59.91,0.00,0,60.36,0.00,0,59.91,0.00,0,60.36,0.00,0,59.66,0.00,0,81.13,0.00,0,51.56,0.00,0,,,-175.8,+20.857
29,PubMed,LOP-GPN,83.94,0.00,0,93.23,0.00,0,0.00,0.00,1,68.96,0.00,0,68.96,0.00,0,68.96,0.00,0,69.10,0.00,0,64.87,0.00,0,64.30,0.00,0,66.28,0.00,0,+48.296,+44.303,-183.5,+56.243,83.99,0.00,1,81.68,0.00,1,77.03,0.00,0,82.71,0.00,0,77.03,0.00,0,75.93,0.00,0,67.74,0.00,0,91.84,0.00,1,80.37,0.00,1,+45.693,+37.657,-160.4,+60.553,83.52,0.00,0,74.83,0.00,0,70.26,0.00,1,70.46,0.00,1,70.26,0.00,1,67.93,0.00,1,55.83,0.00,0,60.86,0.00,0,65.49,0.00,1,,,-195.2,+36.812
30,OGBN\\Arxiv,APPNP,68.76,0.00,1,73.33,0.00,1,0.00,0.00,1,64.39,0.00,0,64.47,0.00,0,64.39,0.00,0,,0.00,0,,0.00,0,,0.00,0,,0.00,0,+28.927,+28.927,,,66.53,0.00,1,27.28,0.00,0,46.98,0.00,0,42.63,0.00,0,46.98,0.00,0,,0.00,0,,0.00,0,,0.00,0,,0.00,0,-12.93,-12.93,,,68.22,0.00,1,65.73,0.00,1,63.98,0.00,0,66.97,0.00,0,63.98,0.00,0,,0.00,0,,0.00,0,,0.00,0,,0.00,0,,,,
31,OGBN\\Arxiv,Matern-GGP,,0.00,0,,0.00,0,,0.00,0,,0.00,0,,0.00,0,,0.00,0,,0.00,0,,0.00,0,,0.00,0,,0.00,0,,,,,,0.00,0,,0.00,0,,0.00,0,,0.00,0,,0.00,0,,0.00,0,,0.00,0,,0.00,0,,0.00,0,,,,,,0.00,0,,0.00,0,,0.00,0,,0.00,0,,0.00,0,,0.00,0,,0.00,0,,0.00,0,,0.00,0,,,,
32,OGBN\\Arxiv,GKDE,56.97,0.00,0,60.04,0.00,0,0.00,0.00,1,69.41,0.00,1,68.69,0.00,1,69.41,0.00,1,66.97,0.00,0,67.46,0.00,1,68.14,0.00,0,66.70,0.00,0,+17.791,+16.112,-186.8,+34.402,56.91,0.00,0,57.59,0.00,1,49.63,0.00,0,49.52,0.00,0,49.63,0.00,0,49.55,0.00,0,49.47,0.00,0,49.46,0.00,0,49.48,0.00,0,-0.431,-0.385,-200.3,-0.939,56.91,0.00,0,57.59,0.00,0,49.63,0.00,0,49.52,0.00,0,49.63,0.00,0,49.55,0.00,0,49.47,0.00,0,49.46,0.00,0,49.48,0.00,0,,,-200.3,-0.939
33,OGBN\\Arxiv,GPN (sym),48.87,0.00,0,55.36,0.00,0,0.00,0.00,1,64.31,0.00,0,67.08,0.00,0,64.31,0.00,0,67.08,0.00,0,66.91,0.00,0,68.79,0.00,1,66.92,0.00,1,+20.547,+20.544,-194.4,+30.225,4.35,0.00,0,4.38,0.00,0,67.63,0.00,1,77.08,0.00,0,67.63,0.00,1,77.45,0.00,1,66.92,0.00,0,69.93,0.00,0,67.14,0.00,0,+2.084,+2.027,-197.8,+21.302,50.35,0.00,0,49.87,0.00,0,50.06,0.00,0,51.15,0.00,0,50.06,0.00,0,51.15,0.00,0,53.41,0.00,0,97.12,0.00,0,50.95,0.00,0,,,-181.2,+8.330
34,OGBN\\Arxiv,GPN (rw),50.52,0.00,0,56.87,0.00,0,0.00,0.00,1,65.63,0.00,0,67.99,0.00,0,65.63,0.00,0,67.99,0.00,0,63.36,0.00,0,66.72,0.00,0,63.37,0.00,0,+21.523,+21.521,-194.4,+40.167,4.35,0.00,0,4.38,0.00,0,66.93,0.00,0,72.27,0.00,0,66.93,0.00,0,72.76,0.00,0,60.35,0.00,0,62.39,0.00,0,60.50,0.00,0,+2.842,+2.742,-197.3,+33.076,50.93,0.00,0,50.66,0.00,0,50.32,0.00,0,51.56,0.00,0,50.32,0.00,0,51.56,0.00,0,52.94,0.00,0,97.25,0.00,1,51.26,0.00,0,,,-181.2,+13.262
35,OGBN\\Arxiv,LOP-GPN,57.74,0.00,0,63.37,0.00,0,0.00,0.00,1,67.00,0.00,0,66.77,0.00,0,67.00,0.00,0,68.12,0.00,1,66.28,0.00,0,67.44,0.00,0,54.58,0.00,0,+16.029,+17.598,-196.0,+7.011,57.49,0.00,0,56.06,0.00,0,67.29,0.00,0,82.46,0.00,1,67.29,0.00,0,73.03,0.00,0,68.39,0.00,1,92.39,0.00,1,87.08,0.00,1,+22.651,+17.534,-183.4,+49.155,57.85,0.00,0,57.93,0.00,0,65.65,0.00,1,70.58,0.00,1,65.65,0.00,1,68.03,0.00,1,63.98,0.00,1,72.86,0.00,0,64.08,0.00,1,,,-194.5,+14.350
\end{filecontents*}
\csvreader[
        column count=99,
        tabular={c r | r rrrrr | r rrrrr | r rrrrr},
        separator=comma,
        table head={%
            & \multicolumn{1}{c}{} &%
            \multicolumn{6}{c}{Leave-out Classes} &%
            \multicolumn{6}{c}{$\mathbf{x}^{(v)} \sim \mathrm{Ber}(0.5)$} &%
            \multicolumn{6}{c}{$\mathbf{x}^{(v)} \sim \mathcal{N}(0,1)$}%
            \\%
            & \multicolumn{1}{r}{} &%
            \multicolumn{1}{c}{\textbf{ID}} & %
            \multicolumn{5}{c}{\textbf{OOD-AUC-ROC}} &%
            \multicolumn{1}{c}{\textbf{ID}} & %
            \multicolumn{5}{c}{\textbf{OOD-AUC-ROC}} &%
            \multicolumn{1}{c}{\textbf{ID}} & %
            \multicolumn{5}{c}{\textbf{OOD-AUC-ROC}} \\
            & \multicolumn{1}{r}{} &%
            \multicolumn{1}{c}{\textbf{Acc}} & %
            \multicolumn{1}{c}{$\mathrm{TU}$} & \multicolumn{1}{c}{$\mathrm{AU}$} & \multicolumn{1}{c}{$\mathrm{EU}$} & \multicolumn{1}{c}{$\mathrm{EU}_{\mathrm{PC}}$} & \multicolumn{1}{c}{$\mathrm{EU}_{\mathrm{SO}}$} &%
            \multicolumn{1}{c}{\textbf{Acc}} & %
            \multicolumn{1}{c}{$\mathrm{TU}$} & \multicolumn{1}{c}{$\mathrm{AU}$} & \multicolumn{1}{c}{$\mathrm{EU}$} & \multicolumn{1}{c}{$\mathrm{EU}_{\mathrm{PC}}$} & \multicolumn{1}{c}{$\mathrm{EU}_{\mathrm{SO}}$} &%
            \multicolumn{1}{c}{\textbf{Acc}} & %
            \multicolumn{1}{c}{$\mathrm{TU}$} & \multicolumn{1}{c}{$\mathrm{AU}$} & \multicolumn{1}{c}{$\mathrm{EU}$} & \multicolumn{1}{c}{$\mathrm{EU}_{\mathrm{PC}}$} & \multicolumn{1}{c}{$\mathrm{EU}_{\mathrm{SO}}$} %
            \\\toprule%
        },
        before reading={\setlength{\tabcolsep}{3pt}},
        table foot=\bottomrule,
        head to column names,
        late after line={\ifthenelse{\equal{\model}{APPNP}\and\not\equal{\id}{0}}{\\\midrule}{\\}},
        filter={\not\(\equal{\dataset}{OGBN\\Arxiv}\and\equal{\model}{Matern-GGP}\)}
    ]{tables/id_ood.csv}{}{%
        \ifthenelse{\equal{\model}{APPNP}}{%
            \ifthenelse{\equal{\dataset}{OGBN\\Arxiv}}{%
                \multirow{5}{*}[0em]{\shortstack{\dataset}}%
            }{\multirow{6}{*}[0em]{\shortstack{\dataset}}}%
        }{} &%
        \textbf{\model} &%
        \rawres{\oodLocIdAcc}{\oodLocIdAccSE}{\oodLocIdAccBest} &%
        \rawres{\oodLocOodTotalEntropy}{\oodLocOodTotalEntropySE}{\oodLocOodTotalEntropyBest} & %
        \rawres{\oodLocOodAleatoricEntropy}{\oodLocOodAleatoricEntropySE}{\oodLocOodAleatoricEntropyBest} & %
        \rawres{\oodLocOodEpistemicEntropyDiff}{\oodLocOodEpistemicEntropyDiffSE}{\oodLocOodEpistemicEntropyDiffBest} & %
        \rawres{\oodLocOodEpistemic}{\oodLocOodEpistemicSE}{\oodLocOodEpistemicBest} & %
        \rawres{\oodLocOodEpistemicEntropy}{\oodLocOodEpistemicEntropySE}{\oodLocOodEpistemicEntropyBest} %
        &%
        \rawres{\oodBerIdAcc}{\oodBerIdAccSE}{\oodBerIdAccBest} &%
        \rawres{\oodBerOodTotalEntropy}{\oodBerOodTotalEntropySE}{\oodBerOodTotalEntropyBest} & %
        \rawres{\oodBerOodAleatoricEntropy}{\oodBerOodAleatoricEntropySE}{\oodBerOodAleatoricEntropyBest} & %
        \rawres{\oodBerOodEpistemicEntropyDiff}{\oodBerOodEpistemicEntropyDiffSE}{\oodBerOodEpistemicEntropyDiffBest} & %
        \rawres{\oodBerOodEpistemic}{\oodBerOodEpistemicSE}{\oodBerOodEpistemicBest} & %
        \rawres{\oodBerOodEpistemicEntropy}{\oodBerOodEpistemicEntropySE}{\oodBerOodEpistemicEntropyBest} %
        &%
        \rawres{\oodNormalIdAcc}{\oodNormalIdAccSE}{\oodNormalIdAccBest} &%
        \rawres{\oodNormalOodTotalEntropy}{\oodNormalOodTotalEntropySE}{\oodNormalOodTotalEntropyBest} & %
        \rawres{\oodNormalOodAleatoricEntropy}{\oodNormalOodAleatoricEntropySE}{\oodNormalOodAleatoricEntropyBest} & %
        \rawres{\oodNormalOodEpistemicEntropyDiff}{\oodNormalOodEpistemicEntropyDiffSE}{\oodNormalOodEpistemicEntropyDiffBest} & %
        \rawres{\oodNormalOodEpistemic}{\oodNormalOodEpistemicSE}{\oodNormalOodEpistemicBest} & %
        \rawres{\oodNormalOodEpistemicEntropy}{\oodNormalOodEpistemicEntropySE}{\oodNormalOodEpistemicEntropyBest} %
    }}
\end{table*}
Next, we evaluate the performance of \ac{lopgpn} and the other models on \acf{ood} detection tasks.
Similar to \citet{stadler2021}, we use three different \ac{ood} detection tasks.
First, we evaluate the models' ability to detect nodes belonging to classes that were not present in the training set.
Second, we randomly drop features from some nodes with probability $0.5$ and evaluate the models' ability to detect those nodes as outliers.
Third, we add Gaussian noise to the features of some nodes, which should then be detected as outliers.
For all detection tasks, we use the entropy-based \ac{tu}, \ac{au} and \ac{eu} measures, as well as $\mathrm{EU}_{\mathrm{PC}}$ and $\mathrm{EU}_{\mathrm{SO}}$, as criteria.
We use the \emph{Area Under Receiving Operator Characteristics Curve} (AUC-ROC) to measure the performance of the models on the \ac{ood} nodes; the performance on the \ac{id} training data is measured via the accuracy.

\Cref{tbl:ood} shows the \ac{ood} detection performance and \ac{id} accuracies.
Overall, \ac{lopgpn} performs very well on the feature dropout and Gaussian noise tasks, outperforming the other models most often.
Looking at the \ac{id} accuracies in the Gaussian noise setting, \ac{lopgpn} is uniquely able to achieve high accuracies on all but the OGBN Arxiv dataset.
On the leave-out-classes detection task, \ac{lopgpn} performs slightly worse but overall still similarly to the other models.

To summarize our experimental results, \ac{lopgpn} was able to achieve strong classification accuracies and meaningful uncertainty estimates, as shown in \cref{fig:accrej} and \cref{tbl:ood}.
This supports our hypothesis that the irreducibility of conflicts assumption made by \ac{gpn} is not always justified in real-world node classification tasks.

\section{Conclusion}\label{sec:conclusion}


In this paper, we proposed a new approach to uncertainty quantification in (semi-supervised) node classification, which is able to represent both aleatoric and epistemic uncertainty.
Broadly speaking, while existing methods realize information dispersion on the level of the data (or pseudo-counts) or the level of aleatoric uncertainty (averaging first-order distributions), our approach makes use of the graph structure to combine information directly on the epistemic level.
To this end, we refer to the established principle of linear opinion pooling and represent epistemic uncertainty in terms of mixtures of Dirichlet distributions.
First experiments on a variety of graph-structured datasets are promising and show the effectiveness of our approach, also compared to state-of-the-art methods used as baselines.

In future work, we plan to study the problem of uncertainty propagation on graphs in more depths and to compare different approaches in a more systematic way.
Intuitively, an optimal approach should find a good compromise between combining information on the aleatoric and the epistemic level, respectively.
However, for now, it is not at all clear how such an approach could be realized.





\bibliography{literature}

\end{document}